%% file: main.tex
\documentclass[sigconf]{acmart}

\usepackage{enumitem}
\usepackage{colortbl}
\usepackage{multirow}
\definecolor{hlca}{HTML}{FFFFE5}
\definecolor{hlcb}{HTML}{ADDD8E}
\definecolor{hlcc}{HTML}{41AB5D}
\definecolor{hlcd}{HTML}{004529}

\copyrightyear{2026}
\acmYear{2026}
\setcopyright{cc}
\setcctype{by}
\acmConference[ICAIF '26]{7th ACM International Conference on AI in Finance}{November 14--17, 2026}{Milan, Italy}
\acmBooktitle{7th ACM International Conference on AI in Finance (ICAIF '26), November 14--17, 2026, Milan, Italy}
\acmDOI{10.1145/3858063.3858793}
\acmISBN{979-8-4007-3087-0/2026/11}

\begin{document}

\title[Time-Series Foundation Models for Financial Crime Detection]{
Leveraging Time-Series Foundation Models for~Financial~Crime~Detection}

\author{Danae Natalie Savvidi}
\affiliation{%
  \institution{Delft University of Technology}
  \city{Delft}
  \country{The Netherlands}}
\email{dsavvidi@tudelft.nl}

\author{\c{C}a\u{g}r\i{} Bilgi}
\affiliation{%
  \institution{Delft University of Technology}
  \city{Delft}
  \country{The Netherlands}}
\email{h.c.bilgi@tudelft.nl}

\author{Kubilay Atasu}
\affiliation{%
  \institution{Delft University of Technology}
  \city{Delft}
  \country{The Netherlands}}
\email{kubilay.atasu@tudelft.nl}

\author{Arie van Deursen}
\affiliation{%
  \institution{Delft University of Technology}
  \city{Delft}
  \country{The Netherlands}}
\email{arie.vandeursen@tudelft.nl}

\renewcommand{\shortauthors}{Savvidi et al.}

\input{sections/1_abstract}



\maketitle

\input{sections/2_introduction}

\input{sections/3_related_work}
\input{sections/4_method}
\input{sections/5_experiments}
\input{sections/6_discussion}
\input{sections/7_conclusion}
\input{sections/8_acknowledgements}

\bibliographystyle{ACM-Reference-Format}
\bibliography{thesis-refs}

\end{document}

%% file: sections/1_abstract.tex
\begin{abstract}
\looseness=-1
Financial crime detection requires identifying illicit activity within
large transaction networks, where suspicious behaviour is rare and
unfolds over time. Existing graph-based methods capture relationships
among accounts but typically underuse the temporal behaviour encoded
in account transaction histories. We introduce
\emph{Temporal-Context Edge Enrichment} (TCEE), a framework that
addresses this gap by summarising each account's transaction history
with a frozen time-series foundation model (TSFM), and enriching each
transaction's representation with the relevant temporal context from
these summaries. These enriched representations can be scored directly
for edge classification, or fed into a graph neural network (GNN) as edge features to
capture broader structural information. Across six AMLWorld datasets,
TCEE improves PR-AUC over the strongest evaluated baseline on five by a
relative $10.8\%$ on average, while training $6.0\times$ faster
end-to-end.
Feeding TCEE's representations into a GNN raises PR-AUC a further $4.8\%$
on the four smaller datasets. On an Ethereum phishing dataset, we show that TSFM-generated account representations fed into a GNN 
are also effective for node classification, improving PR-AUC by $14.5\%$ over the strongest baseline.
As TSFMs continue to advance, our results
suggest they can strengthen financial crime detection by capturing
temporal signals that complement graph-based approaches.
\end{abstract}

%% file: sections/2_introduction.tex
\section{Introduction}
\label{sec:intro}

\looseness=-1
Modern crime is not only physical. Financial crime is carried out through
digital transactions, from money laundering to defrauding account
holders. Money laundering alone is estimated to account for 2 to 5\% of global GDP each
year~\cite{unodc}. Whatever the crime, illicit funds must move, and every
movement is recorded as a transaction between two accounts, so
financial crime unfolds on a transaction network. Such a network is naturally a \emph{multigraph}, with accounts
as nodes, transactions as directed edges, and parallel edges 
representing repeated transactions between account pairs.

\looseness=-1
Finding the few illicit transactions among millions of legitimate
ones is the task of financial crime detection, a legal obligation for
financial institutions~\cite{Oztas2024}. GNNs and graph transformers designed for such multigraphs
define the state of the
art in banking and blockchain settings alike~\cite{Egressy2024,
Lin2024fraudgt, Ding2024diam}, and recent work further extends
them~\cite{Bilgi2024}. We call this family \emph{structural models},
as they focus on the graph structure yet consume
time as a plain edge attribute.

\looseness=-1
Despite advances, detection remains challenging. Illicit
behaviour is vanishingly rare, as few as one transaction in two
thousand~\cite{Altman2023}, and the criminal patterns, sequences of
coordinated transfers, evolve as criminals seek to evade
detection~\cite{Oztas2024}. What must be detected also varies with the
crime.
In phishing, such as on the Ethereum (ETH) blockchain, victims are
tricked into paying an account the attacker controls, so the label is
the account itself and detection is \emph{node
classification}~\cite{wu2020phishers}. Money laundering is different. 
The proceeds can travel through otherwise legitimate
accounts, and what is illicit is each transfer that moves
them, so anti-money laundering (AML) benchmarks pose the problem as
\emph{edge classification}~\cite{Altman2023}. Edge classification is
generally harder, because whether a transaction is suspicious depends
on context. The model must learn not only the characteristics of the
accounts involved but also how they interact.

\looseness=-1
Among these challenges, we target the temporal signal.
Accounts transact in repeating rhythms, salaries arrive and bills are
paid on schedule, and on real bank data these rhythms separate
launderers from legitimate customers~\cite{ketenci2021timefreq}. Two
obstacles stand in the way. Temporal patterns must be learned from
scarce illicit labels, and the rhythms belong to accounts while the
decision falls on a single transaction. The problem is therefore to
enrich each transaction with temporal context drawn from its
structural neighbourhood.

\looseness=-1
Prior work misses this requirement in different ways.
The structural models reduce the temporal signal to a single timestamp
attribute. Temporal graph models are built around time, yet each line of work
reads it narrowly. Snapshot models see only the edges inside a time
step~\cite{weber2019aml}, temporal graph networks compress an account's
past into one memory vector~\cite{TGN2020}, and recency-weighted
message passing reduces timing to one scalar per
neighbour~\cite{Gounoue2025temptrag}.

\looseness=-1
Time-series foundation models (TSFMs) such as
Chronos-2~\cite{Ansari2025} offer a new option. Pretrained on large
corpora, they supply rich temporal representations
without needing any labels. A frozen
TSFM takes each account's history as one series per
feature, the amount, currency and so on, and returns one
embedding per series. Pooling this set into one vector per account,
which we call \emph{node-level fusion}, and using it as a node
feature in a GNN works well for node-level tasks. On a real ETH
phishing graph~\cite{wu2020phishers, xblockEthereum}, this model
outperforms all evaluated baselines. However, it still trails the strongest structural
models on AML edge classification~\cite{Altman2023}, because it hands
every transaction of an account the same pooled vector, while 
edge-level tasks require telling them apart.

\input{figures/per_txn_context}

\looseness=-1
We overcome this limitation by delivering the temporal
context directly to the edge. \emph{Temporal-Context Edge Enrichment} (TCEE) lets each
transaction form a query from its own features and attend over the
frozen embeddings of its source and destination accounts
(Figure~\ref{fig:per_txn_context}), producing a rich edge
representation. Two transactions by the same account can thus retrieve
different contexts: e.g., one drawing on the amount history, and
another on the timing. We further add two \emph{extensions},
components that read the raw transaction features directly and produce further
representation details the temporal context cannot provide.
To our knowledge, TCEE is the first model to query a TSFM from a graph \emph{edge}. The enriched edge representations built by TCEE can be scored directly by an MLP
(\textbf{TCEE-MLP}) or handed to a GNN backbone that can consume edge features (\textbf{TCEE-GNN}).

We make the following main contributions:
\begin{enumerate}[leftmargin=*,itemsep=1pt,topsep=2pt]
\item \looseness=-1 \textbf{Models.} We build, to our knowledge, the first
  TSFM-based detectors of financial crime on transaction multigraphs,
  \textbf{Chronos-MLP} and \textbf{Chronos-GNN}, which use node-level fusion.
\item \textbf{Method.} We show that node-level fusion falls short on edge-level tasks and introduce \textbf{TCEE}, which enables each
  transaction to independently query its temporal context.
\item \textbf{Results.} On AMLWorld, TCEE-MLP with both extensions and no
  backbone improves PR-AUC, the area under the precision-recall
  curve, over each dataset's strongest evaluated baseline on five of
  the six datasets, by a relative $10.8\%$ on average, and trains $6.0\times$
  faster end-to-end on average across all six. A GNN backbone adds a further
  $4.8\%$ on the four datasets trained to convergence. On
  the ETH phishing graph, TCEE improves PR-AUC by $17.8\%$ over the
  strongest baseline. Together, these results make a case for using TSFMs in financial crime detection.
\end{enumerate}

%% file: figures/per_txn_context.tex
\begin{figure}[tb]
\centering
\includegraphics[width=\columnwidth]{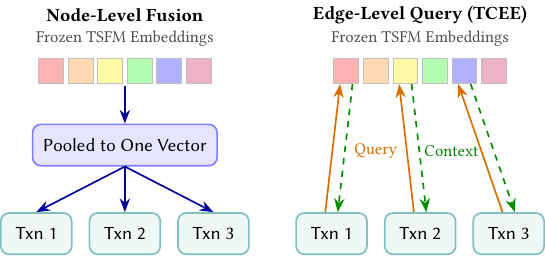}
\caption{Node-level fusion (left) provides the same pooled vector to every transaction of the same account; TCEE (right) enables each transaction to query the context independently.}
\Description{Two panels of the same account. Node-level fusion pools the frozen TSFM
embeddings into one vector shared by all three transactions, while TCEE lets each
transaction query the embedding set and retrieve its own context.}
\label{fig:per_txn_context}
\end{figure}

%% file: sections/3_related_work.tex
\section{Background and Related Work}
\label{sec:background}

\paragraph{Problem} We work on a directed transaction multigraph
$G=(V,E)$. Each node is an account and each edge a transaction
$e=(u,v,t,\mathbf{z}_e)$ from $u$ to $v$ at time $t$ with raw features
$\mathbf{z}_e$. AML is edge classification, where a classifier
$f:e\mapsto\hat{y}_e\in[0,1]$ predicts the illicit label
$y_e\in\{0,1\}$ under severe class imbalance, with illicit transactions
as rare as $0.051\%$ on AMLWorld LI-Medium~\cite{Altman2023}. Phishing
detection is node classification, predicting one label per account.

\paragraph{Feature Engineering} Before GNNs, AML systems paired
hand-engineered features with classifiers. One early approach~\cite{weber2019aml} pairs each transaction's local attributes with one-hop neighbourhood
statistics on the Elliptic Bitcoin graph, and another~\cite{Savage2016mlgroups} extracts community structure and
transaction dynamics per group for a supervised classifier. The Graph Feature
Preprocessor~\cite{Blanusa2024gfp} mines subgraph patterns in real time to
feed gradient-boosted trees, matching or outperforming simple GNN baselines. Such
features are fixed in advance, so patterns they do not express stay
invisible to the classifier.
GNNs instead learn features directly from the graph.

\paragraph{Structural GNNs} The strongest models compete on
\emph{structural} expressiveness, how faithfully the model interprets the graph.
Multi-GNN~\cite{Egressy2024} adds ego IDs, port numbering and reverse message
passing to standard GNNs so they provably recognise directed subgraph patterns
such as cycles. MEGA-GNN~\cite{Bilgi2024} aggregates in
two stages. It first pools the parallel edges between an account pair, then
combines the pooled messages across the account's neighbours.
Multi-FraudGT~\cite{Lin2024fraudgt} uses the augmentations of Multi-GNN and
performs message passing with a graph transformer whose attention is biased
by edge features. On the phishing task, early detectors classify
each account from a random-walk embedding of the transaction
network~\cite{wu2020phishers}. DIAM~\cite{Ding2024diam}, designed for
cryptocurrency multigraphs, compresses each account's transactions
into one node embedding with a GRU. Multi-GNN, MEGA-GNN and
Multi-FraudGT improve how graph structure is read, yet treat a
transaction's \emph{time} as a plain edge attribute. DIAM's GRU models
time explicitly, but it is still trained end-to-end on the
scarce illicit labels rather than pretrained. TCEE supplies what 
both lack, a pretrained temporal representation. It is \emph{orthogonal} to
the structural axis, changing only the edge representation
consumed, so it composes with any of them.

\paragraph{Temporal Modelling} Several lines already model timing. Message
passing can be re-weighted towards recent transactions~\cite{Gounoue2025temptrag}.
Continuous-time dynamic graph networks attach a recurrent memory or a time
encoding to each node~\cite{TGN2020}, discrete-time ones evolve their
weights over graph snapshots~\cite{weber2019aml}, and dedicated detectors
exploit smurfing's transfer timing~\cite{Starnini2021}. The graph models
learn their temporal representation from scratch, and each account is
summarised by a single node embedding that all its transactions share. TCEE
differs on both counts. It reads each account's full history with a frozen
pretrained model, a richer temporal signal, and it
delivers that signal to the individual transaction, the unit AML
classifies, rather than diluting it in node aggregation.

\paragraph{Frozen Time-Series Encoders}
TSFMs are neural networks pretrained on large, diverse collections of time series, so they learn general temporal patterns and forecast new
series with no task-specific training~\cite{Ansari2025}. Most split each series into short,
fixed-length, non-overlapping segments called patches, and treat them as input
tokens~\cite{nie2023patchtst}. Architectures range from encoder-only
models such as Chronos-2~\cite{Ansari2025} and Moirai~\cite{Woo2024} to
decoder-only forecasters such as TimesFM~\cite{Das2024}, and from
univariate to multivariate models that jointly model related series.
\input{figures/encoding_pipeline}
What matters for TCEE is a set of embeddings per account to attend
over, not one pooled vector. Any TSFM with an encoder provides this,
so the approach follows TSFM advances rather than depending on one
model. 
We use Chronos-2, a
leading multivariate, encoder-only forecaster, kept frozen. Each
account $a$ has $F$ parallel feature series, one per transaction feature
(Section~\ref{sec:method_encoding}). Chronos-2 encodes the $F$ series of an account jointly. Its group attention performs
in-context learning, meaning that the representation of every feature
series is conditioned on the account's remaining $F-1$ series without
any weight updates. The encoder
reads each in patches of $16$ transactions, and returns the tensor
$\mathbf{E}_a\in\mathbb{R}^{F\times P_a\times768}$ (Figure~\ref{fig:encoding}),
one $768$-dimensional embedding per series and patch, with $P_a$ the
account's patch count. TCEE queries these embeddings with each transaction.
Graph models that condition
attention on edge features~\cite{diao2023relational} or
message-pass on the line graph~\cite{Poon2025linemvgnn} operate over
representations learned inside the graph, and work that brings pretrained
encoders to GNNs fuses them at the \emph{node} level~\cite{shao2022step}. To
our knowledge, TCEE is the first to query a frozen encoder from a
graph \emph{edge}, and to bring a TSFM to financial crime detection.

%% file: figures/encoding_pipeline.tex
\begin{figure}[tb]
\centering
\includegraphics[width=\columnwidth]{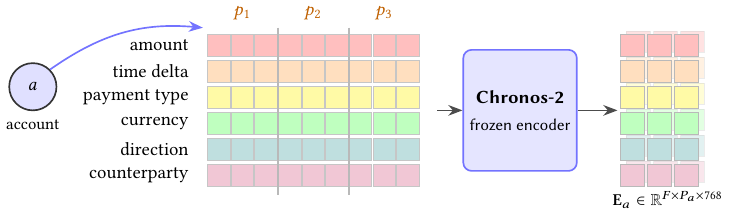}
\caption{Chronos-2, frozen, reads each account's history in
patches of $16$ transactions and emits one $768$-d embedding per feature and
patch.}
\Description{An account's transaction history, split into six feature streams and
patches of sixteen transactions, passes through the frozen Chronos-2 encoder and
emerges as the embedding tensor $\mathbf{E}_a$ of shape $F$ by $P_a$ by 768.}
\label{fig:encoding}
\end{figure}

%% file: sections/4_method.tex
\section{Method}
\label{sec:method}

This section covers the encoding of transaction histories as time series,
node-level fusion, TCEE and the cost analysis. All attention below is
single-head scaled dot-product
attention~\cite{Vaswani2017},
$\mathrm{Attn}(\mathbf{q},\mathbf{K},\mathbf{V})
= \mathrm{softmax}\big(\mathbf{q}\mathbf{K}^{\top}\!/\sqrt{d}\,\big)\mathbf{V}$
for a query $\mathbf{q}$, keys $\mathbf{K}\in\mathbb{R}^{n\times d}$ and
values $\mathbf{V}$.

\subsection{Transaction History as a Time Series}
\label{sec:method_encoding}
A transaction history is not the time series a TSFM expects, for two
reasons. First, transactions are discrete events at irregular intervals, while
time-series models often assume regular sampling. Second, each transaction involves
two accounts, so a history is part of a graph, not a self-contained series.
Our construction addresses both.

We build one sequence per account, containing every transaction it sends or
receives, ordered by time. Each transaction is one step of the sequence.
For AMLWorld each step carries $F{=}6$ features, so the history becomes six
parallel series, one per feature (Figure~\ref{fig:encoding}). The
\emph{amount}, \emph{payment type} and \emph{currency} describe the
transaction itself, and the \emph{direction} flag marks it as incoming or
outgoing for the account. The remaining two
series close the two gaps above. The \emph{time delta} $\Delta t_i = t_i - t_{i-1}$, in minutes, 
with $t_i$ the time of the account's $i$-th
transaction and $\Delta t_1 = 0$, closes the first. Chronos-2 treats its input as equally
spaced, so the delta is the simplest way to expose the spacing between
steps. The \emph{counterparty index} closes the second. It numbers the
account's counterparties in the order they first appear, so a returning partner
keeps its number. Through this series the encoder sees the account's one-hop
neighbourhood, so the representation it returns is not purely temporal. We
ablate the counterparty index (Table~\ref{tab:ablation}) to measure how much of the
signal is temporal. Categorical and boolean features become
integers, as the official implementation
accepts only real-valued series.

Before training, Chronos-2 encodes each account's $F$ series jointly into the
embedding tensor $\mathbf{E}_a\in\mathbb{R}^{F\times P_a\times768}$
(Figure~\ref{fig:encoding}). We average $\mathbf{E}_a$ over its patches,
$\bar{\mathbf{E}}_a=\tfrac{1}{P_a}\sum_{p=1}^{P_a}\mathbf{E}_a^{(p)}
\in\mathbb{R}^{F\times768}$, with $\mathbf{E}_a^{(p)}$ the $p$-th patch, 
leaving one embedding per feature for each account. We ablate the averaging
(Table~\ref{tab:ablation}).

\subsection{Node-Level Fusion}
\label{sec:method_baselines}

\looseness=-1
A simple way to use the frozen embeddings is \emph{node-level fusion}, pooling
them into the same vector per account. Two such models measure what the frozen
signal is worth on its own. Each model pools the account's
feature embeddings $\bar{\mathbf{E}}_a$ into one vector
$\mathbf{h}_a\in\mathbb{R}^{d_{\text{proj}}}$. By default we pool with
attention. A linear map per feature projects $\bar{\mathbf{E}}_a$ to
$\tilde{\mathbf{E}}_a\in\mathbb{R}^{F\times d_{\text{proj}}}$, and a learned query
$\mathbf{q}_0$ attends over the projections,
\begin{equation}
\mathbf{h}_a=\mathbf{W}_o\,\mathrm{Attn}\big(\mathbf{q}_0,\tilde{\mathbf{E}}_a,\tilde{\mathbf{E}}_a\big),
\label{eq:pool}
\end{equation}
with $\mathbf{W}_o$ a learned output map. The attention lets each account
weight its most informative series, and replacing its weights with a uniform
mean gives a plain-mean variant we ablate (Table~\ref{tab:ablation}).
Chronos-MLP classifies an edge from its two accounts,
\begin{equation}
\hat{y}_e=\mathrm{MLP}\big([\,\mathbf{h}_u\,\|\,\mathbf{h}_v\,\|\,
  \mathbf{h}_u{\odot}\mathbf{h}_v\,\|\,|\mathbf{h}_u{-}\mathbf{h}_v|
  \,\|\,\mathbf{z}_e\,]\big),
\label{eq:nodebaseline}
\end{equation}
with $\|$ denoting concatenation, $\odot$ the Hadamard product and
$\mathbf{z}_e$ the edge's raw features. The
product and absolute difference are standard pair-matching
features~\cite{conneau2017infersent} that show the classifier how the two accounts
align and differ. Chronos-GNN instead uses the same
$\mathbf{h}_a$ as account $a$'s initial node feature in the GNN backbone
(Section~\ref{sec:method_readout}).


\subsection{TCEE: Temporal-Context Edge Enrichment}
\label{sec:method_tcee}

TCEE lets each transaction attend over the frozen embeddings of its two accounts,
$\bar{\mathbf{E}}_u$ and $\bar{\mathbf{E}}_v$, with a query built from the transaction's
own features. Figure~\ref{fig:arch}a shows the architecture.

\input{figures/arch}

\begin{samepage}
\subsubsection{The Edge-Conditioned Query}

\paragraph{Keys and Values.} For each account $a$ we project its averaged
feature embeddings $\bar{\mathbf{E}}_a$ (Section~\ref{sec:method_encoding}) to $F$
keys and values,
\end{samepage}
\begin{equation}
\mathbf{K}_a = (\bar{\mathbf{E}}_a+\mathbf{T})\mathbf{W}_K,\qquad
\mathbf{V}_a = (\bar{\mathbf{E}}_a+\mathbf{T})\mathbf{W}_V,\qquad
\mathbf{K}_a,\mathbf{V}_a\in\mathbb{R}^{F\times d_{\text{attn}}},
\label{eq:kv}
\end{equation}
where $\mathbf{W}_K,\mathbf{W}_V\in\mathbb{R}^{768\times d_{\text{attn}}}$ are shared across all
accounts and $\mathbf{T}\in\mathbb{R}^{F\times768}$ is a learned per-feature type
embedding that tags each of the $F$ feature series before projection.

\paragraph{The Query Is the Transaction.} The same transaction looks different
to its two accounts. Its direction, for example, is outgoing for the sender
and incoming for the receiver. We write $\mathbf{z}_e$ from each account's perspective, a sender view
$\mathbf{z}_e^{u}$ and a receiver view $\mathbf{z}_e^{v}$, and encode both with one
shared MLP $\phi$ into account queries $\mathbf{q}_u=\phi(\mathbf{z}_e^{u})$ and
$\mathbf{q}_v=\phi(\mathbf{z}_e^{v})$ in $\mathbb{R}^{d_{\text{attn}}}$. Each query attends over
its own account's keys (Figure~\ref{fig:arch}a),
\begin{equation}
\mathbf{ctx}_u = \mathrm{Attn}(\mathbf{q}_u,\mathbf{K}_u,\mathbf{V}_u),\qquad
\mathbf{ctx}_v = \mathrm{Attn}(\mathbf{q}_v,\mathbf{K}_v,\mathbf{V}_v).
\label{eq:ctx}
\end{equation}
The keys $\mathbf{K}_a$ and values $\mathbf{V}_a$ are fixed per account while
the query changes with every transaction, so two transfers of the same
account can weight its feature embeddings differently.

\subsubsection{Local Context}
\label{sec:method_branches}

The context retrieved in Eq.~\eqref{eq:ctx} is rich but coarse, a
summary of whole account histories. It is
blind to the specific exchange between $u$ and $v$ and to the timing right around
the transaction. We add two extensions, one for each blind spot.
\emph{Pair pooling} aggregates the transactions between $u$ and $v$, and
\emph{neighbourhood attention} reads the transactions nearest to $e$ in
each account's history (Figure~\ref{fig:branches}b). Both adapt prior GNN
components.

\paragraph{Pair Pooling.} Let $M_{uv}=\{e'\in E: e'\text{ connects }u\text{ and
}v\}$ be the transactions between the pair. We summarise them with PNA~\cite{Corso2020}, which applies four
aggregators $\alpha$ to the multiset $M$, rescales each result by three degree
scalers $s$, and concatenates the twelve outputs,
\begin{equation}
\begin{gathered}
\mathrm{PNA}(M)={\textstyle\big\Vert_{\alpha,\,s}}\;s\,\alpha(M),\qquad
\alpha\in\{\mathrm{mean},\mathrm{std},\mathrm{max},\mathrm{min}\},\\
s\in\{1,\ \lambda,\ \lambda^{-1}\},\qquad
\lambda=\log(1{+}|M|)\,/\,\mu,
\end{gathered}
\label{eq:pna}
\end{equation}
where $\mu$, the mean of $\log(1+\mathrm{deg})$ over the training set,
normalises $\lambda$ to average one. Applied to $M_{uv}$, with each
transaction first encoded from its $F$ raw features by a small MLP, this gives
$\mathbf{b}^{\text{pair}}_{uv}$. This is the intra-pair pooling that
MEGA-GNN~\cite{Bilgi2024} performs as the first stage of its aggregation.
The account-level summary includes the pair's
transactions but mixes them into the account's full history, while the pair
pooling extension isolates them.

\paragraph{Neighbourhood Attention.} The transactions just before and after the
target, in each account's own history, carry local signal the pooled account
embedding averages away. For each account $a$ we attend over
$\delta_w(a,e)$, a window of $2w$ transactions around $e$ in $a$'s history. Writing $t_e$ for the time of the target edge, the window holds the $w$ most 
recent transactions before $t_e$ and the first $w$ from $t_e$ onward, the
target itself included. If one side holds fewer than $w$ transactions, the window
extends on the other side, and a history of at most $2w$ transactions is
taken whole. The keys $\mathbf{K}_{\delta}$ and values $\mathbf{V}_{\delta}$
are projected from the $F$ raw features of the transactions in $\delta_w(a,e)$. Attention is permutation invariant while the
window is an ordered sequence, so we append two features to each
transaction as positional encoding relative to $e$, the time difference from
$t_e$ and a flag, 1 for the entries after $t_e$ and 0 otherwise. The account query
$\mathbf{q}_a$, projected by
$\mathbf{W}_n$, attends over the window,
\begin{equation}
\mathbf{n}_a = \mathrm{Attn}\big(\mathbf{W}_n\mathbf{q}_a,\,
   \mathbf{K}_{\delta},\,\mathbf{V}_{\delta}\big),
\label{eq:nodeattn}
\end{equation}
and the outputs for the two accounts concatenate into
$\mathbf{b}^{\text{nbr}}_{uv}=[\mathbf{n}_u\,\|\,\mathbf{n}_v]$. This is one hop on the
line graph of $G$, whose nodes are the edges of $G$, adjacent when they share an
account. Prior AML work processes this line-graph view with message
passing~\cite{Poon2025linemvgnn}. We use attention instead,
restricted to each account's temporal window. The fixed
window keeps it a constant-cost local operation, without the neighbourhood
sampling message passing needs. The two extension outputs concatenate into
$\mathbf{b}_{uv}=[\mathbf{b}^{\text{pair}}_{uv}\,\|\,\mathbf{b}^{\text{nbr}}_{uv}]$.

\subsubsection{The Enriched Edge}

We now fuse the account queries $\mathbf{q}_u,\mathbf{q}_v$, the contexts
$\mathbf{ctx}_u,\mathbf{ctx}_v$ retrieved from the edge's two accounts, and
the extension outputs $\mathbf{b}_{uv}$ into one edge representation. The
account queries are already encoded by $\phi$, so we combine them with one
linear map and layer normalisation,
$\mathbf{q}_{uv}=\mathrm{LN}\big(\mathbf{W}_f[\mathbf{q}_u\,\|\,\mathbf{q}_v]\big)$.
The enriched edge representation is then (Figure~\ref{fig:arch}a)
\begin{equation}
\begin{aligned}
\mathbf{e}^{\text{rich}}_{uv} = \mathrm{LN}(\mathbf{W}_r\mathbf{q}_{uv})
  + \mathrm{MLP}\big([\,&\mathbf{q}_{uv}\,\|\,\mathbf{ctx}_u\,\|\,\mathbf{ctx}_v\,\|\\
  &\mathbf{ctx}_u{\odot}\mathbf{ctx}_v\,\|\,|\mathbf{ctx}_u{-}\mathbf{ctx}_v|
  \,\|\,\mathbf{b}_{uv}\,]\big),
\end{aligned}
\label{eq:enrich}
\end{equation}
where $\mathbf{W}_r$ is a learned linear map and $\mathbf{b}_{uv}$ the optional
extension outputs of Section~\ref{sec:method_branches}. The skip keeps the
transaction's own features, encoded in $\mathbf{q}_{uv}$, on a direct path
to the output. The contexts enter
through the same matching features as at the node
(Eq.~\eqref{eq:nodebaseline}), $\mathbf{ctx}_u{\odot}\mathbf{ctx}_v$ and
$|\mathbf{ctx}_u{-}\mathbf{ctx}_v|$.

\subsubsection{Classifier and Backbone}
\label{sec:method_readout}

TCEE changes only the edge representation, so $\mathbf{e}^{\text{rich}}_{uv}$
drops into any downstream model. TCEE-MLP scores it directly with an MLP,
while TCEE-GNN passes it to a GNN backbone. For the GNN backbone we use
PNA~\cite{Corso2020}, the aggregator of the best performing baseline
variants on most AMLWorld datasets~\cite{Bilgi2024, Egressy2024}.
We name the resulting models TCEE-PNA and Chronos-PNA, which feed the backbone
differently.
TCEE-PNA injects $\mathbf{e}^{\text{rich}}_{uv}$ at the edge,
\begin{equation}
\mathbf{h}^{(0)}_a=\mathbf{0},\qquad
\mathbf{e}^{(0)}_{uv}=\mathbf{W}_E\,\mathbf{e}^{\text{rich}}_{uv},
\label{eq:tceegnn}
\end{equation}
with $\mathbf{W}_E$ a learned map. Chronos-PNA injects $\mathbf{h}_a$ at the
node,
\begin{equation}
\mathbf{h}^{(0)}_a=\mathbf{W}_C\,\mathbf{h}_a,\qquad
\mathbf{e}^{(0)}_{uv}=\mathbf{W}_E\,\mathbf{z}_e,
\label{eq:chronosgnn}
\end{equation}
while the edges carry only projected raw features. Every model is trained
end-to-end with the class-weighted cross-entropy
\begin{equation}
\mathcal{L}=-\sum_{e}\,w_{y_e}\big[\,y_e\log\hat{y}_e
   + (1-y_e)\log(1-\hat{y}_e)\,\big],
\label{eq:loss}
\end{equation}
where $\hat{y}_e$ is the illicit probability and the class weight
$w_{y_e}$ up-weights the rare illicit class ($y_e{=}1$), as in the baselines.

\subsubsection{Node Classification}
\label{sec:method_node}
On a node task the label belongs to an account, not a transaction.
TCEE still enriches every edge (Eq.~\eqref{eq:tceegnn}), the GNN backbone's message passing pools
an account's incident enriched edges into its node state 
$\mathbf{h}^{(L)}_a$ after $L$ layers, and a per-node head classifies the
account, $\hat{y}_a=\mathrm{MLP}\big(\mathbf{h}^{(L)}_a\big)$, instead of
the edge head. Training uses the loss of Eq.~\eqref{eq:loss} with the sum
over accounts.

\subsection{Computational Resources}
\label{sec:method_complexity}

Cost matters in deployment, with banks screening millions of transactions
and retraining as criminal behaviour changes.
We derive compute and storage costs and compare with the structural models.

\subsubsection{Computational Complexity}

The frozen encoder runs once per account and split before training. Encoding account
$a$ is one transformer forward pass over its $F\times P_a$ grid of patch
tokens, so building the store costs
$O\big(F\sum_{a}(P_a^{2}\,c+P_a\,c^{2})\big)$, with $c{=}768$ the encoder
width. Every model and epoch reuses this store, so the cost is one-time.
Training cost depends on what a minibatch holds, a list of edges or a
sampled subgraph. Let $d$ be the largest hidden width.

Chronos-MLP and TCEE-MLP train on minibatches of edges. For every edge, both
project the two endpoints' stored embeddings, $F$ vectors of width $c$ per
account, down to width $d$, costing $F\,c\,d$ per account. The projection
weights change at every optimiser step, so the result cannot be precomputed
and reused. Chronos-MLP reduces the $F$ projected vectors to one account
vector (Eq.~\eqref{eq:pool}) and scores the edge with the MLP head, so an
epoch over $|E|$ edges costs
\begin{equation}
O\big(|E|\,[\underbrace{F\,c\,d}_{\text{pool, Eq.~\eqref{eq:pool}}}
+\underbrace{d^{2}}_{\text{head, Eq.~\eqref{eq:nodebaseline}}}]\big).
\label{eq:mlpcost}
\end{equation}
TCEE-MLP pays the same projection to form its keys and values, then adds
three computations. The attention of Eq.~\eqref{eq:ctx} scores the
transaction's query against each account's $F$ keys, pair pooling aggregates
the $|M_{uv}|$ transactions between $u$ and $v$ (Eq.~\eqref{eq:pna}), and
neighbourhood attention reads a window of $2w$ transactions per
account (Eq.~\eqref{eq:nodeattn}). Both extensions first encode each 
transaction from its $F$ raw features, so one edge costs
\begin{equation}
O\big(\underbrace{F\,c\,d}_{\text{keys and values}}
+\underbrace{F\,d}_{\text{account attn.}}
+\underbrace{|M_{uv}|\,F\,d}_{\text{pair pool.}}
+\underbrace{2w\,F\,d}_{\text{nbr.\ attn.}}
+\underbrace{d^{2}}_{\text{MLPs}}\big).
\label{eq:edgecost}
\end{equation}
Only $|M_{uv}|$ varies across edges, so an epoch costs
$O\big(|E|\,[F(c+1+\overline{|M|}+2w)\,d+d^{2}]\big)$, with $\overline{|M|}$
the mean of $|M_{uv}|$ over the training edges. Two things follow. First, no
term depends on account degree, account count or a multi-hop neighbourhood,
so both models are linear in $|E|$. Second, both epochs are dominated by the
shared $F\,c\,d$ projection, since $1{+}\overline{|M|}{+}2w\ll c$ and
$d\ll F\,c$, which is why the two are close in wall-clock (Table~\ref{tab:cost}).

Chronos-PNA and TCEE-PNA train on sampled subgraphs with edges $E_{s}$ and
nodes $V_{s}$ instead. A subgraph holds far more edges than nodes, so the $F\,c\,d$ projection of an account's stored embeddings is paid once per node rather than once per edge,
$O\big(|V_{s}|\,F\,c\,d\big)$ per step. The saving does not show
because the $L$-layer PNA backbone ($L{=}2$) applies
Eq.~\eqref{eq:pna} to every sampled neighbour and its
$O\big(L\,(|E_{s}|{+}|V_{s}|)\,d^{2}\big)$ dominates the step. The
baselines MEGA-GNN~\cite{Bilgi2024}, Multi-GNN~\cite{Egressy2024} and
Multi-FraudGT~\cite{Lin2024fraudgt} pay this same
backbone cost. They score an edge by message passing over a sampled
neighbourhood that grows with fanout and depth.

\subsubsection{Storage}

Each embedding tensor $\mathbf{E}_a$ holds $768\cdot F\cdot P_a$ values, so
at \texttt{float16} (2 bytes) the store over $N$ accounts is
\begin{equation}
S \;=\; 2\cdot768\cdot F\sum_{a=1}^{N} P_a \quad\text{bytes}.
\label{eq:storage}
\end{equation}
An account seen in $n_a$ transactions unfolds into $P_a=\lceil n_a/16\rceil$ size-$16$
patches, capped at an $8{,}192$-transaction context ($512$ patches), so $\sum_a P_a$
grows with the transaction count. Averaging over patches (Section~\ref{sec:method_encoding}) sets
$P_a{=}1$, collapsing this to $\bar{\mathbf{E}}_a\in\mathbb{R}^{F\times768}$ and the
store to $2\cdot768\cdot F\cdot N$ bytes, a flat $9$~KB per account ($F{=}6$).
Table~\ref{tab:datasets} reports the full graph storage cost, which recurs per
split. We keep the store GPU-resident for faster training, and when it
exceeds the GPU's VRAM (medium and large) we hold only the training split
resident and memory-map the rest.

\begin{table}[tb]
\centering
\caption{Dataset statistics.}
\label{tab:datasets}
\Description{Node, edge, embedding size, illicit rate, time span and split proportions for the six AMLWorld graphs and the Ethereum phishing graph.}
\setlength{\tabcolsep}{4pt}
\resizebox{\columnwidth}{!}{%
\begin{tabular}{lrrrrrr}
\toprule
Dataset & \# Nodes & \# Edges & Emb.\ (GB) & Illicit & Days & Split [\%] \\
\midrule
HI-Small  & 515{,}088     & 5{,}078{,}345   & 4.7  & 0.102\% & 10   & 64/19/17 \\
LI-Small  & 705{,}907     & 6{,}924{,}049   & 6.5  & 0.051\% & 10   & 64/19/17 \\
HI-Medium & 2{,}077{,}023 & 31{,}898{,}238  & 19.1 & 0.110\% & 16   & 61/17/22 \\
LI-Medium & 2{,}032{,}095 & 31{,}251{,}483  & 18.7 & 0.051\% & 16   & 61/17/22 \\
HI-Large  & 2{,}116{,}168 & 179{,}702{,}229 & 19.5 & 0.124\% & 97   & 60/20/20 \\
LI-Large  & 2{,}070{,}980 & 176{,}066{,}557 & 19.1 & 0.057\% & 97   & 60/20/20 \\
\midrule
ETH       & 2{,}973{,}489 & 13{,}551{,}303  & 18.3 & 0.04\%  & 1261 & 65/15/20 \\
\bottomrule
\end{tabular}%
}
\end{table}

%% file: figures/arch.tex
\begin{figure*}[t]
\centering
\begin{minipage}[b]{0.72\textwidth}
\centering
\includegraphics[width=\linewidth]{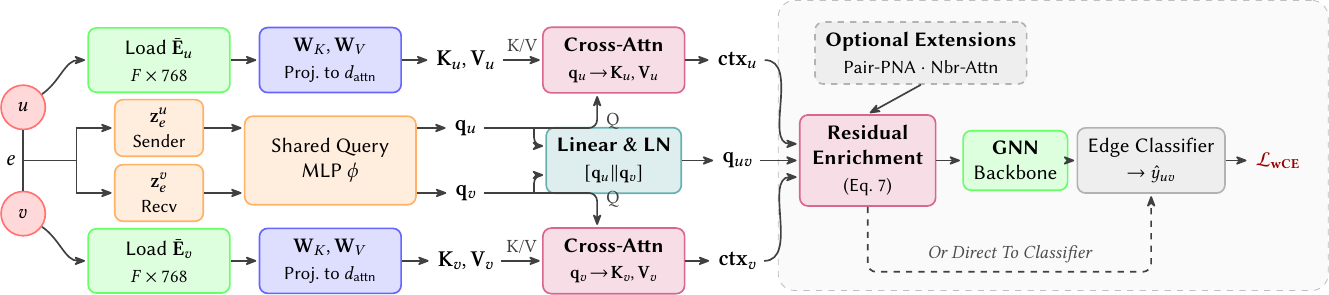}\\
{\small (a) Architecture}
\end{minipage}\hfill
\begin{minipage}[b]{0.28\textwidth}
\centering
\includegraphics[width=\linewidth]{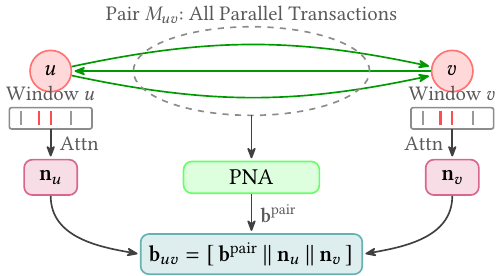}\\
{\small (b) Local-context extensions}
\end{minipage}

\caption{(a) TCEE architecture. The transaction edge $e=(u,v)$ takes each account's frozen
Chronos-2 embedding (green) as keys and values, forms a query from the two transaction
views through a shared MLP $\phi$, and cross-attends for $\mathbf{ctx}_u,\mathbf{ctx}_v$.
Residual enrichment fuses the queries, contexts and optional
extensions into $\mathbf{e}^{\text{rich}}_{uv}$, scored directly or after a GNN
backbone. The temporal signal enters at the edge, not the node. (b) The extensions. Pair pooling
aggregates the parallel transactions with PNA, and
neighbourhood attention reads each account's window, giving
$\mathbf{b}_{uv}$.}
\Description{Left, a block diagram of TCEE. The frozen embeddings of both
endpoints supply keys and values, the two transaction views form queries
through a shared MLP, and cross-attention returns the contexts that
residual enrichment fuses into an enriched edge, scored either directly or
after a GNN backbone. Right, the two local-context extensions. PNA pools all
parallel transactions between the accounts while attention over each
account's window gives the neighbourhood outputs, and the three
concatenate into the extension feature.}
\label{fig:arch}
\label{fig:branches}
\end{figure*}

%% file: sections/5_experiments.tex
\begin{table*}[t]
\centering
\caption{Test PR-AUC and minority-class F1 (\%) on the six AMLWorld datasets,
mean $\pm$ std over 5 seeds.}
\label{tab:main}
\Description{Test PR-AUC and minority-class F1 for every model on the six AMLWorld datasets.}
\footnotesize
\setlength{\tabcolsep}{2pt}
\renewcommand{\arraystretch}{0.9}
\resizebox{\textwidth}{!}{%
\begin{tabular}{l cc cc cc cc cc cc}
\toprule
& \multicolumn{2}{c}{\textbf{HI-Small}} & \multicolumn{2}{c}{\textbf{LI-Small}}
& \multicolumn{2}{c}{\textbf{HI-Medium}} & \multicolumn{2}{c}{\textbf{LI-Medium}}
& \multicolumn{2}{c}{\textbf{HI-Large}} & \multicolumn{2}{c}{\textbf{LI-Large}} \\
\cmidrule(lr){2-3}\cmidrule(lr){4-5}\cmidrule(lr){6-7}\cmidrule(lr){8-9}\cmidrule(lr){10-11}\cmidrule(lr){12-13}
\textbf{Model} & PR-AUC & F1 & PR-AUC & F1 & PR-AUC & F1 & PR-AUC & F1 & PR-AUC & F1 & PR-AUC & F1 \\
\midrule
GINe & \cellcolor{hlcc!80!hlcb}$49.94_{\pm4.02}$ & \cellcolor{hlcc!94!hlcb}$53.77_{\pm3.39}$ & \cellcolor{hlcb!29!hlca}$16.77_{\pm4.27}$ & \cellcolor{hlcb!40!hlca}$21.25_{\pm4.86}$ & \cellcolor{hlcc!47!hlcb}$44.55_{\pm4.53}$ & \cellcolor{hlcb!85!hlca}$34.20_{\pm9.98}$ & \cellcolor{hlcb!25!hlca}$14.95_{\pm3.42}$ & \cellcolor{hlcb!25!hlca}$14.93_{\pm3.45}$ & \cellcolor{hlcc!10!hlcb}$38.34_{\pm7.04}$ & \cellcolor{hlcc!27!hlcb}$41.19_{\pm5.65}$ & \cellcolor{hlcb!6!hlca}$6.50_{\pm1.77}$ & \cellcolor{hlcb!3!hlca}$5.10_{\pm5.30}$\\
PNAe & \cellcolor{hlcd!15!hlcc}$59.40_{\pm3.18}$ & \cellcolor{hlcd!21!hlcc}$60.93_{\pm3.35}$ & \cellcolor{hlcb!42!hlca}$22.29_{\pm1.40}$ & \cellcolor{hlcb!68!hlca}$31.26_{\pm2.80}$ & \cellcolor{hlcd!12!hlcc}$58.44_{\pm3.96}$ & \cellcolor{hlcd!10!hlcc}$58.12_{\pm4.65}$ & \cellcolor{hlcb!39!hlca}$21.04_{\pm2.33}$ & \cellcolor{hlcb!53!hlca}$26.99_{\pm3.73}$ & \cellcolor{hlcd!17!hlcc}$59.90_{\pm0.49}$ & \cellcolor{hlcd!23!hlcc}$61.37_{\pm0.98}$ & \cellcolor{hlcb!39!hlca}$20.81_{\pm1.68}$ & \cellcolor{hlcb!50!hlca}$25.48_{\pm3.45}$\\
\midrule
Multi-GNN~\cite{Egressy2024} & \cellcolor{hlcd!51!hlcc}\textcolor{white}{$69.05_{\pm1.22}$} & \cellcolor{hlcc!76!hlcb}$49.28_{\pm4.11}$ & \cellcolor{hlcb!62!hlca}$30.30_{\pm2.39}$ & \cellcolor{hlcb!87!hlca}$34.56_{\pm1.41}$ & \cellcolor{hlcd!74!hlcc}\textcolor{white}{$74.97_{\pm0.60}$} & \cellcolor{hlcd!67!hlcc}\textcolor{white}{$73.26_{\pm1.31}$} & \cellcolor{hlcb!95!hlca}$35.86_{\pm0.10}$ & \cellcolor{hlcc!43!hlcb}$43.78_{\pm0.59}$ & \cellcolor{hlcd!68!hlcc}\textcolor{white}{$73.57_{\pm0.46}$} & \cellcolor{hlcd!64!hlcc}\textcolor{white}{$72.36_{\pm1.29}$} & \cellcolor{hlcb!54!hlca}$27.44_{\pm1.48}$ & \cellcolor{hlcb!81!hlca}$33.54_{\pm2.29}$\\
MEGA-GNN~\cite{Bilgi2024} & \cellcolor{hlcd!71!hlcc}\textcolor{white}{$74.28_{\pm2.56}$} & \cellcolor{hlcd!69!hlcc}\textcolor{white}{$73.63_{\pm2.45}$} & \cellcolor{hlcc!10!hlcb}$38.41_{\pm1.08}$ & \cellcolor{hlcc!53!hlcb}$45.49_{\pm2.49}$ & \cellcolor{hlcd!86!hlcc}\textcolor{white}{$78.32_{\pm0.82}$} & \cellcolor{hlcd!85!hlcc}\textcolor{white}{$78.00_{\pm0.23}$} & \cellcolor{hlcc!18!hlcb}$39.60_{\pm3.38}$ & \cellcolor{hlcc!73!hlcb}$48.75_{\pm1.65}$ & \cellcolor{hlcd!83!hlcc}\textcolor{white}{$77.40_{\pm0.21}$} & \cellcolor{hlcd!80!hlcc}\textcolor{white}{$76.68_{\pm0.57}$} & \cellcolor{hlcb!59!hlca}$29.36_{\pm1.01}$ & \cellcolor{hlcb!88!hlca}$34.62_{\pm1.77}$\\
Multi-FraudGT~\cite{Lin2024fraudgt} & \cellcolor{hlcd!75!hlcc}\textcolor{white}{$75.46_{\pm0.39}$} & \cellcolor{hlcd!76!hlcc}\textcolor{white}{$75.68_{\pm0.78}$} & \cellcolor{hlcc!6!hlcb}$37.72_{\pm0.75}$ & \cellcolor{hlcc!51!hlcb}$45.20_{\pm1.56}$ & \cellcolor{hlcd!73!hlcc}\textcolor{white}{$74.88_{\pm0.36}$} & \cellcolor{hlcd!77!hlcc}\textcolor{white}{$75.97_{\pm0.20}$} & \cellcolor{hlcb!97!hlca}$36.15_{\pm0.57}$ & \cellcolor{hlcc!48!hlcb}$44.66_{\pm0.66}$ & \cellcolor{hlcd!65!hlcc}\textcolor{white}{$72.62_{\pm0.37}$} & \cellcolor{hlcd!68!hlcc}\textcolor{white}{$73.48_{\pm0.68}$} & \cellcolor{hlcb!58!hlca}$29.16_{\pm0.10}$ & \cellcolor{hlcb!93!hlca}$35.52_{\pm0.17}$\\
\midrule
Chronos-MLP & \cellcolor{hlcd!13!hlcc}$58.92_{\pm3.39}$ & \cellcolor{hlcd!18!hlcc}$60.04_{\pm1.48}$ & \cellcolor{hlcb!37!hlca}$20.16_{\pm3.53}$ & \cellcolor{hlcb!46!hlca}$24.05_{\pm7.55}$ & \cellcolor{hlcd!37!hlcc}\textcolor{white}{$65.15_{\pm1.02}$} & \cellcolor{hlcd!31!hlcc}$63.50_{\pm4.34}$ & \cellcolor{hlcb!54!hlca}$27.30_{\pm1.41}$ & \cellcolor{hlcb!85!hlca}$34.13_{\pm1.17}$ & \cellcolor{hlcd!36!hlcc}\textcolor{white}{$64.89_{\pm1.08}$} & \cellcolor{hlcd!13!hlcc}$58.83_{\pm1.59}$ & \cellcolor{hlcb!46!hlca}$23.79_{\pm0.84}$ & \cellcolor{hlcb!57!hlca}$28.51_{\pm0.47}$\\
Chronos-PNA & \cellcolor{hlcd!36!hlcc}\textcolor{white}{$64.89_{\pm2.17}$} & \cellcolor{hlcd!26!hlcc}$62.22_{\pm2.17}$ & \cellcolor{hlcb!48!hlca}$24.92_{\pm3.31}$ & \cellcolor{hlcb!74!hlca}$32.39_{\pm2.44}$ & \cellcolor{hlcd!48!hlcc}\textcolor{white}{$68.20_{\pm1.88}$} & \cellcolor{hlcd!46!hlcc}\textcolor{white}{$67.73_{\pm2.98}$} & \cellcolor{hlcb!65!hlca}$30.89_{\pm1.08}$ & \cellcolor{hlcc!10!hlcb}$38.25_{\pm0.98}$ & \cellcolor{hlcd!47!hlcc}\textcolor{white}{$67.85_{\pm1.12}$$^\ast$} & \cellcolor{hlcd!44!hlcc}\textcolor{white}{$67.16_{\pm1.44}$$^\ast$} & \cellcolor{hlcb!53!hlca}$26.75_{\pm1.04}$$^\ast$ & \cellcolor{hlcb!63!hlca}$30.45_{\pm2.13}$$^\ast$\\
\midrule
TCEE-MLP & \cellcolor{hlcd!52!hlcc}\textcolor{white}{$69.24_{\pm0.61}$} & \cellcolor{hlcd!50!hlcc}\textcolor{white}{$68.66_{\pm2.02}$} & \cellcolor{hlcb!69!hlca}$31.45_{\pm0.93}$ & \cellcolor{hlcc!3!hlcb}$37.18_{\pm1.05}$ & \cellcolor{hlcd!73!hlcc}\textcolor{white}{$74.70_{\pm0.53}$} & \cellcolor{hlcd!71!hlcc}\textcolor{white}{$74.25_{\pm0.66}$} & \cellcolor{hlcb!92!hlca}$35.25_{\pm0.36}$ & \cellcolor{hlcc!32!hlcb}$41.92_{\pm0.41}$ & \cellcolor{hlcd!76!hlcc}\textcolor{white}{$75.50_{\pm0.96}$} & \cellcolor{hlcd!71!hlcc}\textcolor{white}{$74.34_{\pm0.90}$} & \cellcolor{hlcb!76!hlca}$32.67_{\pm0.77}$ & \cellcolor{hlcc!9!hlcb}$38.13_{\pm0.95}$\\
\quad $+$ Pair-PNA & \cellcolor{hlcd!74!hlcc}\textcolor{white}{$75.00_{\pm0.60}$} & \cellcolor{hlcd!70!hlcc}\textcolor{white}{$73.95_{\pm1.85}$} & \cellcolor{hlcb!84!hlca}$34.04_{\pm1.15}$ & \cellcolor{hlcc!24!hlcb}$40.64_{\pm1.15}$ & \cellcolor{hlcd!89!hlcc}\textcolor{white}{$79.11_{\pm0.10}$} & \cellcolor{hlcd!88!hlcc}\textcolor{white}{$78.75_{\pm0.50}$} & \cellcolor{hlcc!8!hlcb}$37.98_{\pm0.91}$ & \cellcolor{hlcc!44!hlcb}$44.02_{\pm0.71}$ & \cellcolor{hlcd!94!hlcc}\textcolor{white}{$80.31_{\pm0.37}$} & \cellcolor{hlcd!91!hlcc}\textcolor{white}{$79.65_{\pm0.18}$} & \cellcolor{hlcc!9!hlcb}$38.18_{\pm0.54}$ & \cellcolor{hlcc!39!hlcb}$43.21_{\pm0.42}$\\
\quad $+$ Nbr-Attn & \cellcolor{hlcd!76!hlcc}\textcolor{white}{$75.51_{\pm0.87}$} & \cellcolor{hlcd!64!hlcc}\textcolor{white}{$72.41_{\pm2.46}$} & \cellcolor{hlcb!90!hlca}$34.99_{\pm1.28}$ & \cellcolor{hlcc!9!hlcb}$38.22_{\pm1.93}$ & \cellcolor{hlcd!89!hlcc}\textcolor{white}{$78.97_{\pm0.31}$} & \cellcolor{hlcd!85!hlcc}\textcolor{white}{$77.91_{\pm0.58}$} & \cellcolor{hlcc!11!hlcb}$38.42_{\pm0.87}$ & \cellcolor{hlcc!47!hlcb}$44.54_{\pm1.30}$ & \cellcolor{hlcd!93!hlcc}\textcolor{white}{$80.26_{\pm0.07}$} & \cellcolor{hlcd!85!hlcc}\textcolor{white}{$77.93_{\pm0.31}$} & \cellcolor{hlcc!11!hlcb}$38.48_{\pm0.18}$ & \cellcolor{hlcc!38!hlcb}$43.08_{\pm0.37}$\\
\quad $+$ Pair-PNA $+$ Nbr-Attn & \cellcolor{hlcd!86!hlcc}\textcolor{white}{$78.26_{\pm1.13}$} & \cellcolor{hlcd!80!hlcc}\textcolor{white}{$76.75_{\pm1.26}$} & \cellcolor{hlcb!84!hlca}$34.06_{\pm1.05}$ & \cellcolor{hlcc!23!hlcb}$40.51_{\pm2.41}$ & \cellcolor{hlcd!96!hlcc}\textcolor{white}{$81.00_{\pm0.03}$} & \cellcolor{hlcd!96!hlcc}\textcolor{white}{$80.97_{\pm0.47}$} & \cellcolor{hlcc!27!hlcb}$41.12_{\pm0.43}$ & \cellcolor{hlcc!59!hlcb}$46.54_{\pm0.96}$ & \cellcolor{hlcd!99!hlcc}\textcolor{white}{$81.61_{\pm0.09}$} & \cellcolor{hlcd!98!hlcc}\textcolor{white}{$81.35_{\pm0.16}$} & \cellcolor{hlcc!22!hlcb}$40.39_{\pm0.11}$ & \cellcolor{hlcc!52!hlcb}$45.25_{\pm0.22}$\\
\midrule
TCEE-PNA & \cellcolor{hlcd!71!hlcc}\textcolor{white}{$74.34_{\pm1.25}$} & \cellcolor{hlcd!67!hlcc}\textcolor{white}{$73.17_{\pm1.36}$} & \cellcolor{hlcb!69!hlca}$31.48_{\pm2.25}$ & \cellcolor{hlcb!97!hlca}$36.18_{\pm1.41}$ & \cellcolor{hlcd!87!hlcc}\textcolor{white}{$78.56_{\pm0.62}$} & \cellcolor{hlcd!82!hlcc}\textcolor{white}{$77.33_{\pm0.59}$} & \cellcolor{hlcc!23!hlcb}$40.57_{\pm0.29}$ & \cellcolor{hlcc!54!hlcb}$45.69_{\pm1.70}$ & \cellcolor{hlcd!86!hlcc}\textcolor{white}{$78.24_{\pm0.26}$$^\ast$} & \cellcolor{hlcd!74!hlcc}\textcolor{white}{$75.03_{\pm0.52}$$^\ast$} & \cellcolor{hlcb!98!hlca}$36.32_{\pm0.72}$$^\ast$ & \cellcolor{hlcc!18!hlcb}$39.65_{\pm0.78}$$^\ast$\\
\quad $+$ Pair-PNA & \cellcolor{hlcd!89!hlcc}\textcolor{white}{$79.00_{\pm0.66}$} & \cellcolor{hlcd!81!hlcc}\textcolor{white}{$76.81_{\pm1.40}$} & \cellcolor{hlcb!96!hlca}$36.08_{\pm0.83}$ & \cellcolor{hlcc!26!hlcb}$40.92_{\pm2.24}$ & \cellcolor{hlcd!99!hlcc}\textcolor{white}{$81.61_{\pm0.13}$} & \cellcolor{hlcd!97!hlcc}\textcolor{white}{$81.33_{\pm0.25}$} & \cellcolor{hlcc!42!hlcb}$43.61_{\pm0.41}$ & \cellcolor{hlcc!77!hlcb}$49.54_{\pm0.33}$ & \cellcolor{hlcd!94!hlcc}\textcolor{white}{$80.30_{\pm0.50}$$^\ast$} & \cellcolor{hlcd!82!hlcc}\textcolor{white}{$77.30_{\pm0.89}$$^\ast$} & \cellcolor{hlcc!12!hlcb}$38.69_{\pm0.90}$$^\ast$ & \cellcolor{hlcc!34!hlcb}$42.25_{\pm1.13}$$^\ast$\\
\quad $+$ Pair-PNA $+$ Nbr-Attn & \cellcolor{hlcd!92!hlcc}\textcolor{white}{$79.94_{\pm0.79}$} & \cellcolor{hlcd!85!hlcc}\textcolor{white}{$77.88_{\pm1.34}$} & \cellcolor{hlcc!4!hlcb}$37.32_{\pm1.47}$ & \cellcolor{hlcc!36!hlcb}$42.69_{\pm2.76}$ & \cellcolor{hlcd!99!hlcc}\textcolor{white}{$81.81_{\pm0.22}$} & \cellcolor{hlcd!98!hlcc}\textcolor{white}{$81.55_{\pm0.44}$} & \cellcolor{hlcc!42!hlcb}$43.74_{\pm0.99}$ & \cellcolor{hlcc!79!hlcb}$49.91_{\pm0.71}$ & \cellcolor{hlcd!94!hlcc}\textcolor{white}{$80.50_{\pm0.37}$$^\ast$} & \cellcolor{hlcd!83!hlcc}\textcolor{white}{$77.53_{\pm0.45}$$^\ast$} & \cellcolor{hlcc!14!hlcb}$38.96_{\pm0.47}$$^\ast$ & \cellcolor{hlcc!39!hlcb}$43.16_{\pm0.95}$$^\ast$\\
\bottomrule
\end{tabular}}

\smallskip
{\footnotesize
$^\ast$ Due to computational limitations, those runs were not trained to convergence
and Pair-PNA was capped at the $64$ most recent transactions per pair,
so they understate the true performance.}
\end{table*}

\begin{table}[tb]
\centering
\caption{Extraction and end-to-end wall-clock cost in minutes. End-to-end
includes extraction.}
\label{tab:cost}
\Description{One-time extraction cost and end-to-end wall-clock training time in minutes for each model on each dataset.}
\footnotesize
\setlength{\tabcolsep}{1.5pt}
\renewcommand{\arraystretch}{0.85}
\resizebox{\columnwidth}{!}{%
\begin{tabular}{@{}c@{\hspace{2pt}}l cccccc@{}}
\toprule
& \textbf{Model} & \textbf{HI-S} & \textbf{LI-S} & \textbf{HI-M} & \textbf{LI-M} & \textbf{HI-L} & \textbf{LI-L}\\
\midrule
& Extraction & 39 & 58 & 177 & 181 & 523 & 509\\
\midrule
\multirow{7}{*}{\rotatebox[origin=c]{90}{\scriptsize\textit{End-to-end}}}
& Chronos-MLP & \cellcolor{hlcd!100!hlcc}\textcolor{white}{$43_{\pm2}$} & \cellcolor{hlcd!64!hlcc}\textcolor{white}{$67_{\pm4}$} & \cellcolor{hlcc!83!hlcb}$206_{\pm8}$ & \cellcolor{hlcc!83!hlcb}$202_{\pm3}$ & \cellcolor{hlcb!96!hlca}$837_{\pm69}$ & \cellcolor{hlcb!99!hlca}$780_{\pm67}$\\
& Chronos-PNA & \cellcolor{hlcd!43!hlcc}\textcolor{white}{$89_{\pm11}$} & \cellcolor{hlcd!10!hlcc}$141_{\pm20}$ & \cellcolor{hlcb!99!hlca}$783_{\pm47}$ & \cellcolor{hlcb!99!hlca}$766_{\pm80}$ & \cellcolor{hlcb!18!hlca}$4592_{\pm472}$ & \cellcolor{hlcb!21!hlca}$4414_{\pm220}$\\
\cmidrule(lr){2-8}
& TCEE-MLP & \cellcolor{hlcd!97!hlcc}\textcolor{white}{$44_{\pm1}$} & \cellcolor{hlcd!67!hlcc}\textcolor{white}{$65_{\pm2}$} & \cellcolor{hlcc!77!hlcb}$224_{\pm7}$ & \cellcolor{hlcc!77!hlcb}$223_{\pm11}$ & \cellcolor{hlcb!96!hlca}$843_{\pm49}$ & \cellcolor{hlcb!96!hlca}$810_{\pm62}$\\
& \quad $+$ Pair-PNA & \cellcolor{hlcd!94!hlcc}\textcolor{white}{$47_{\pm2}$} & \cellcolor{hlcd!64!hlcc}\textcolor{white}{$68_{\pm3}$} & \cellcolor{hlcc!68!hlcb}$259_{\pm14}$ & \cellcolor{hlcc!65!hlcb}$269_{\pm16}$ & \cellcolor{hlcb!69!hlca}$1344_{\pm135}$ & \cellcolor{hlcb!66!hlca}$1453_{\pm118}$\\
& \quad $+$ Pair-PNA $+$ Nbr-Attn & \cellcolor{hlcd!88!hlcc}\textcolor{white}{$50_{\pm3}$} & \cellcolor{hlcd!61!hlcc}\textcolor{white}{$70_{\pm3}$} & \cellcolor{hlcc!68!hlcb}$257_{\pm4}$ & \cellcolor{hlcc!65!hlcb}$261_{\pm12}$ & \cellcolor{hlcb!66!hlca}$1420_{\pm184}$ & \cellcolor{hlcb!66!hlca}$1488_{\pm168}$\\
& TCEE-PNA & \cellcolor{hlcd!25!hlcc}$115_{\pm25}$ & \cellcolor{hlcc!98!hlcb}$165_{\pm45}$ & \cellcolor{hlcb!78!hlca}$1181_{\pm83}$ & \cellcolor{hlcb!75!hlca}$1202_{\pm183}$ & \cellcolor{hlcb!9!hlca}$6639_{\pm260}$ & \cellcolor{hlcb!12!hlca}$6032_{\pm260}$\\
& \quad $+$ Pair-PNA $+$ Nbr-Attn & \cellcolor{hlcc!74!hlcb}$232_{\pm22}$ & \cellcolor{hlcc!71!hlcb}$246_{\pm47}$ & \cellcolor{hlcb!42!hlca}$2517_{\pm295}$ & \cellcolor{hlcb!39!hlca}$2597_{\pm191}$ & \cellcolor{hlcb!0!hlca}$11041_{\pm113}$ & \cellcolor{hlcb!0!hlca}$10706_{\pm325}$\\
\midrule
\multirow{3}{*}{\rotatebox[origin=c]{90}{\scriptsize\textit{Baselines}}}
& Multi-GNN & \cellcolor{hlcc!47!hlcb}$354_{\pm107}$ & \cellcolor{hlcc!53!hlcb}$321_{\pm15}$ & \cellcolor{hlcb!30!hlca}$3294_{\pm143}$ & \cellcolor{hlcb!42!hlca}$2497_{\pm187}$ & \cellcolor{hlcb!12!hlca}$5880_{\pm1030}$ & \cellcolor{hlcb!24!hlca}$3897_{\pm45}$\\
& MEGA-GNN & \cellcolor{hlcd!4!hlcc}$152_{\pm87}$ & \cellcolor{hlcd!1!hlcc}$160_{\pm36}$ & \cellcolor{hlcb!66!hlca}$1482_{\pm1169}$ & \cellcolor{hlcb!66!hlca}$1432_{\pm951}$ & \cellcolor{hlcb!15!hlca}$5291_{\pm650}$ & \cellcolor{hlcb!6!hlca}$7422_{\pm1474}$\\
& Multi-FraudGT & \cellcolor{hlcc!8!hlcb}$680_{\pm241}$ & \cellcolor{hlcb!87!hlca}$958_{\pm52}$ & \cellcolor{hlcb!36!hlca}$2852_{\pm38}$ & \cellcolor{hlcb!39!hlca}$2648_{\pm226}$ & \cellcolor{hlcb!6!hlca}$6879_{\pm716}$ & \cellcolor{hlcb!3!hlca}$8231_{\pm261}$\\
\bottomrule
\end{tabular}%
}
\end{table}

\section{Experiments}
\label{sec:experiments}
\begin{samepage}
Our experiments answer four research questions, the first three on AML
edge classification.
\begin{enumerate}[label=\textbf{RQ\arabic*.},leftmargin=*,itemsep=1pt,topsep=2pt]
\item How effective (PR-AUC and minority-class F1) is TCEE?
\item What does TCEE cost to train (time-to-best)?
\item How much can the frozen TSFM contribute?
\item How well does TCEE transfer to phishing detection?
\end{enumerate}
\end{samepage}

\subsection{Experimental Setup}
\label{sec:setup}

\subsubsection{Datasets}
We use AMLWorld~\cite{Altman2023}, an agent-based transaction simulator
with complete labels. It comes in higher- and lower-illicit-ratio regimes (HI, LI)
at small, medium and large sizes. We further evaluate on the ETH phishing
graph~\cite{wu2020phishers, xblockEthereum}, a node- rather than
edge-classification task. Table~\ref{tab:datasets} summarises all graphs.

\subsubsection{Baselines}
\label{sec:exp_baselines}
The structural floor is GINe and PNAe, the standard GIN~\cite{Xu2019gin}
and PNA~\cite{Corso2020} message-passing
architectures extended with edge features, as configured in the AML
baselines~\cite{Egressy2024}. Strong prior models are
MEGA-GNN~\cite{Bilgi2024}, Multi-FraudGT~\cite{Lin2024fraudgt} and
Multi-GNN~\cite{Egressy2024}. MEGA-GNN and
Multi-GNN each come with a GIN, PNA and GenAgg aggregator. We run all
and report per dataset the aggregator with the best test PR-AUC\footnote{PNA in most
cases, GenAgg for MEGA-GNN on HI-Small and LI-Large and for Multi-GNN on
LI-Small.}, so every reported baseline entry upper-bounds its variants. On the
ETH graph, we also compare with the state-of-the-art DIAM~\cite{Ding2024diam}.

\subsubsection{Evaluation}
PR-AUC is the primary metric, as it summarises overall performance under
heavy class imbalance. We also report minority-class F1, a thresholded
metric, for comparability with prior AML models~\cite{Egressy2024, Bilgi2024}.
As in those models, its threshold is fixed at $0.5$ on the predicted
illicit probability.
For a controlled comparison our GINe,
PNAe and the backbones of Chronos-PNA and TCEE-PNA use the
hyperparameters of Multi-GNN and MEGA-GNN~\cite{Egressy2024, Bilgi2024}.
We tune our models' learning rate, attention window size and the widths
$d_{\text{proj}}$ and $d_{\text{attn}}$ on validation PR-AUC. Following the
baselines~\cite{Bilgi2024,Egressy2024,Lin2024fraudgt}, which message-pass over
all edges in a split, embeddings are computed over each
full split. Each run
trains on an NVIDIA A40 GPU with early stopping within a fixed epoch budget, so quality and cost stay
comparable.
We report each test metric at the checkpoint that maximises its validation
score, averaged over 5 seeds. Cost is \emph{time-to-best}, the wall-clock
time to the best validation PR-AUC. Reported percentage gains between models are relative.

\subsection{Experimental Results}
\label{sec:results}

\subsubsection{Effectiveness (RQ1)}
\label{sec:results_rq1}
We discuss Table~\ref{tab:main} in three steps.

First, the frozen Chronos-2 embeddings are strong on their own.
Chronos-MLP, with no message passing, beats the GINe structural floor
on all AML datasets and PNAe on the medium and large graphs, reaching
$65.15$ PR-AUC on HI-Medium against PNAe's $58.44$. A PNA backbone
lifts Chronos-PNA higher, yet both Chronos models trail Multi-GNN and
MEGA-GNN, as node-level fusion hands every transaction of an account the
same vector.

Second, building the representation per transaction adds more. TCEE-MLP
consumes the same embeddings as Chronos-MLP, raising PR-AUC on every
dataset, $28.5\%$ on average.


\looseness=-1
Third, the two extensions help. With both on, the full TCEE-MLP
beats the strongest baseline of each dataset in PR-AUC on five of
the six, with margins averaging $10.8\%$ and reaching $37.6\%$ on LI-Large. It also leads in
minority-class F1 on four of those five, trailing
MEGA-GNN on LI-Medium. Adding the PNA
backbone raises PR-AUC a further $4.8\%$ 
on the four datasets trained to convergence, and $2.4\%$ on average 
across all six. TCEE-MLP trails
only on LI-Small, which holds the fewest illicit transactions of any AML
dataset. There the full TCEE-PNA narrows the gap to within one
standard deviation, $37.3$ PR-AUC against MEGA-GNN's $38.4$.

\subsubsection{Efficiency (RQ2)}
Table~\ref{tab:cost} reports
time-to-best including the one-time extraction all models
reuse. The plain TCEE-MLP reaches its
best score $7.7\times$ faster end-to-end on average than the strongest
baseline, and the full variant $6.0\times$ faster. TCEE-MLP adds $8.0$ to $11.3$ 
PR-AUC points over Chronos-MLP
at almost no extra cost.

\looseness=-1
Figure~\ref{fig:cost_quality} plots quality against cost, with better
models at the top left. The Chronos
models are cheap but score lowest, the structural baselines score well
but cost the most, and the TCEE-MLP variants are cheap and score well,
holding every dataset's Pareto
front. The GNN backbone improves PR-AUC but trains
$7.0\times$ longer end-to-end than the full TCEE-MLP on the four
datasets trained to convergence, and $7.1\times$ across all six datasets.

\subsubsection{Ablation Studies (RQ3)}

\begin{table}[tb]
\centering
\caption{Ablations (PR-AUC \%). $F$ feature series, $P$ patches.}
\label{tab:ablation}
\Description{PR-AUC on HI-Small and LI-Small when the counterparty index, the account pooling, the attention inputs and the Chronos context are each removed or varied.}
\footnotesize
\setlength{\tabcolsep}{4pt}
\renewcommand{\arraystretch}{0.82}
\resizebox{\columnwidth}{!}{%
\begin{tabular}{ll cc}
\toprule
\textbf{Ablation} & \textbf{Setting} & \textbf{HI-Small} & \textbf{LI-Small}\\
\midrule
\multirow{4}{*}{Counterparty}
 & Chronos-MLP, $-$CP & \cellcolor{hlcc!77!hlcb}$54.69_{\pm4.15}$ & \cellcolor{hlcb!0!hlca}$18.80_{\pm3.16}$\\
 & Chronos-MLP, $+$CP & \cellcolor{hlcd!10!hlcc}$58.92_{\pm3.39}$ & \cellcolor{hlcb!3!hlca}$20.16_{\pm3.53}$\\
 & TCEE-MLP, $-$CP    & \cellcolor{hlcd!43!hlcc}\textcolor{white}{$63.20_{\pm2.84}$} & \cellcolor{hlcb!24!hlca}$27.64_{\pm0.83}$\\
 & TCEE-MLP, $+$CP (default) & \cellcolor{hlcd!94!hlcc}\textcolor{white}{$69.24_{\pm0.61}$} & \cellcolor{hlcb!39!hlca}$31.45_{\pm0.93}$\\
\midrule
\multirow{2}{*}{$\mathbf{h}_a$ pool}
 & Chronos-MLP, mean over $F$            & \cellcolor{hlcd!7!hlcc}$58.74_{\pm4.32}$ & \cellcolor{hlcb!3!hlca}$19.98_{\pm2.29}$\\
 & Chronos-MLP, attn.\ over $F$ (default) & \cellcolor{hlcd!10!hlcc}$58.92_{\pm3.39}$ & \cellcolor{hlcb!3!hlca}$20.16_{\pm3.53}$\\
\midrule
\multirow{4}{*}{\shortstack[l]{Pooling\\(TCEE-MLP)}}
 & Mean over $F$ and $P$                    & \cellcolor{hlcd!76!hlcc}\textcolor{white}{$67.24_{\pm1.58}$} & \cellcolor{hlcb!36!hlca}$30.84_{\pm0.62}$\\
 & Attn.\ over $F$, PNA over $P$            & \cellcolor{hlcd!85!hlcc}\textcolor{white}{$68.13_{\pm1.51}$} & \cellcolor{hlcb!36!hlca}$30.54_{\pm1.37}$\\
 & Attn.\ over $F$ and $P$                  & \cellcolor{hlcd!94!hlcc}\textcolor{white}{$69.13_{\pm0.78}$} & \cellcolor{hlcb!33!hlca}$29.74_{\pm1.38}$\\
 & Attn.\ over $F$, mean over $P$ (default) & \cellcolor{hlcd!94!hlcc}\textcolor{white}{$69.24_{\pm0.61}$} & \cellcolor{hlcb!39!hlca}$31.45_{\pm0.93}$\\
\midrule
\multirow{2}{*}{\shortstack[l]{Context\\(TCEE-MLP)}}
 & $-$Chronos                     & \cellcolor{hlcc!80!hlcb}$55.01_{\pm0.87}$ & \cellcolor{hlcb!12!hlca}$23.63_{\pm0.81}$\\
 & $+$Chronos (default)           & \cellcolor{hlcd!94!hlcc}\textcolor{white}{$69.24_{\pm0.61}$} & \cellcolor{hlcb!39!hlca}$31.45_{\pm0.93}$\\
\bottomrule
\end{tabular}%
}
\end{table}
We ablate the frozen TSFM's input and
pooling (Table~\ref{tab:ablation}). On HI-Small and LI-Small, removing the
counterparty index costs Chronos-MLP $4.2$ and $1.4$ PR-AUC points and
TCEE-MLP $6.0$ and $3.8$, so structure is informative.

Attending over an account's $F$ feature embeddings helps only when the query
is the transaction. Chronos-MLP's attention over $F$ matches a plain mean,
while in TCEE-MLP it adds $2.0$ and $0.6$ points.

\looseness=-1
Along the patch axis, PNA and attention over both $F$ and $P$ stay within
two points of the plain mean and win nowhere, as about $63\%$ of accounts fit in a single
$16$-transaction patch. Longer medium and large histories might reward richer
pooling, but the per-patch store outgrows the GPU
(Section~\ref{sec:method_complexity}). Finally, we remove the Chronos
context from TCEE-MLP entirely. With no
keys or values in Eq.~\eqref{eq:kv} there is nothing to attend over, and the
model reduces to an MLP over the fused transaction queries $\mathbf{q}_{uv}$.
This costs $14.2$ and $7.8$ points, the largest drop of any ablation.

\subsubsection{Phishing Detection (RQ4)}
\label{sec:results_rq4}

\begin{table}[tb]
\centering
\caption{ETH phishing node classification, PR-AUC and minority-class F1 (\%).}
\label{tab:eth}
\Description{PR-AUC and minority-class F1 for node classification on the Ethereum phishing graph.}
\footnotesize
\setlength{\tabcolsep}{4pt}
\renewcommand{\arraystretch}{0.9}
\begin{tabular}{l cc}
\toprule
\textbf{Model} & \textbf{PR-AUC} & \textbf{F1} \\
\midrule
Multi-GNN~\cite{Egressy2024} & \cellcolor{hlcc!80!hlcb}$61.2_{\pm0.6}$ & \cellcolor{hlcd!7!hlcc}$65.0_{\pm1.1}$\\
MEGA-GNN~\cite{Bilgi2024} & \cellcolor{hlcc!74!hlcb}$60.5_{\pm4.5}$ & \cellcolor{hlcc!92!hlcb}$62.7_{\pm5.8}$\\
DIAM~\cite{Ding2024diam} & \cellcolor{hlcc!38!hlcb}$55.3_{\pm1.8}$ & \cellcolor{hlcd!4!hlcc}$64.6_{\pm1.4}$\\
\midrule
Chronos-MLP & \cellcolor{hlcb!12!hlca}$31.4_{\pm0.7}$ & \cellcolor{hlcb!0!hlca}$26.0_{\pm2.0}$\\
Chronos-PNA & \cellcolor{hlcd!49!hlcc}\textcolor{white}{$70.1_{\pm0.9}$} & \cellcolor{hlcd!97!hlcc}\textcolor{white}{$75.5_{\pm1.3}$}\\
TCEE-PNA & \cellcolor{hlcd!55!hlcc}\textcolor{white}{$70.9_{\pm1.1}$} & \cellcolor{hlcd!94!hlcc}\textcolor{white}{$75.4_{\pm0.8}$}\\
\quad $+$ Pair-PNA $+$ Nbr-Attn & \cellcolor{hlcd!67!hlcc}\textcolor{white}{$72.1_{\pm0.9}$} & \cellcolor{hlcd!82!hlcc}\textcolor{white}{$73.8_{\pm1.2}$}\\
\bottomrule
\end{tabular}
\end{table}

\input{figures/cost_quality}

For the node task, we evaluate phishing detection on the ETH graph
against Multi-GNN~\cite{Egressy2024},
MEGA-GNN~\cite{Bilgi2024} and DIAM~\cite{Ding2024diam}. TCEE-MLP scores
transactions rather than accounts, so we instead use TCEE-PNA with a
per-node head (Section~\ref{sec:method_node}). Chronos-MLP classifies
each account from its pooled $\mathbf{h}_a$ alone and reaches only
$31.4$ PR-AUC, possibly because ETH carries four features against
AMLWorld's six. With a PNA backbone, the
same embeddings excel, Chronos-PNA and TCEE-PNA scoring $14.5\%$ and $15.9\%$
above the strongest baseline in PR-AUC (Table~\ref{tab:eth}).
The two match because message passing pools an
account's edges into one node state, averaging out the distinctions
TCEE builds. On this task, edge-level querying on its own shows no
measurable benefit over node-level fusion. The two extensions 
do add a little on PR-AUC, with the full TCEE-PNA 
reaching $72.1$ PR-AUC, $17.8\%$
above the strongest baseline, but give slightly worse F1 than plain TCEE-PNA.

%% file: figures/cost_quality.tex
\begin{figure}[tb]
\centering
\includegraphics[width=0.95\columnwidth]{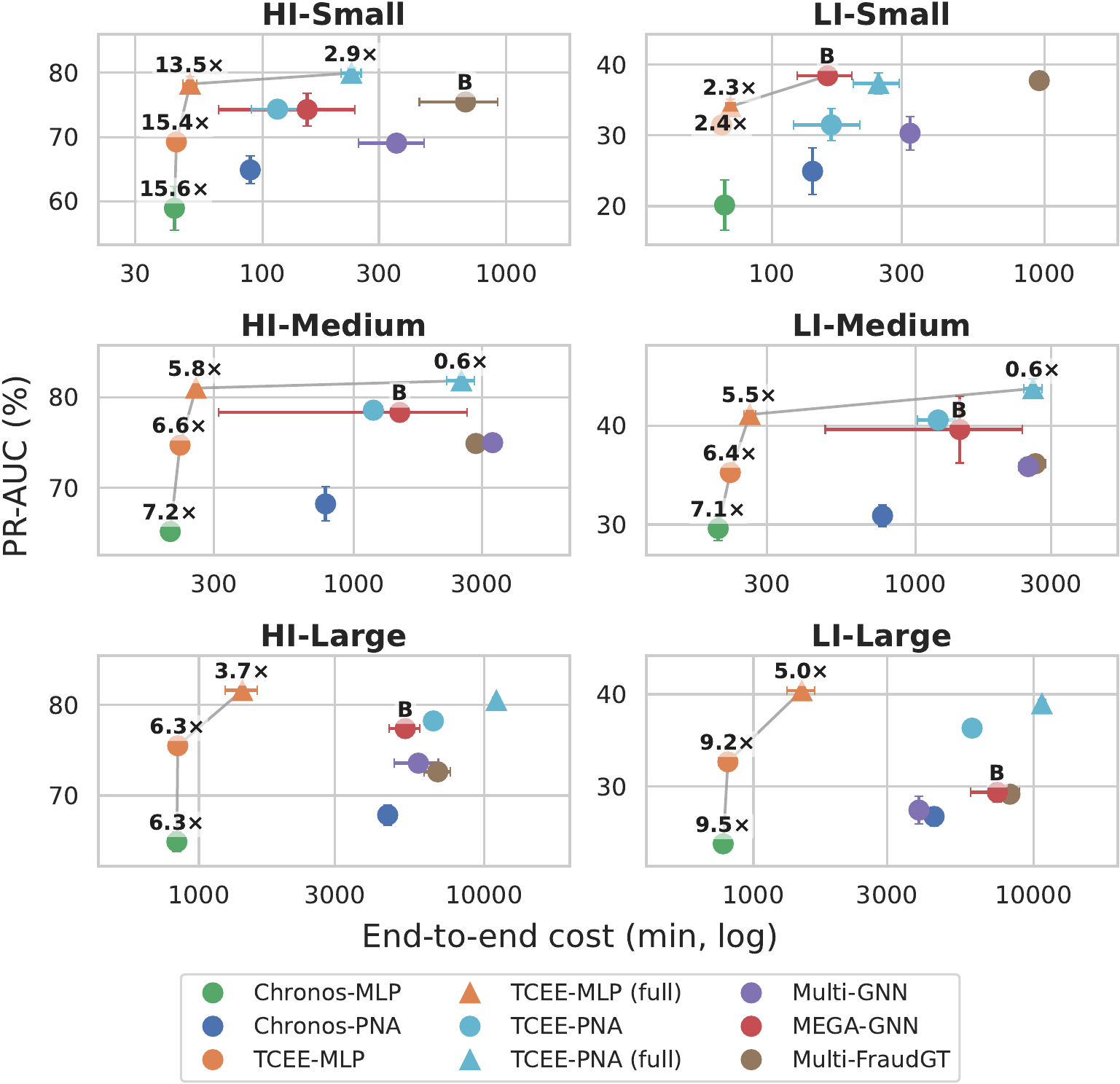}
\caption{Test PR-AUC (\%) against end-to-end cost (log minutes). Top left is
better, the grey line the Pareto front, and $\times N$ each model's
speed-up over the strongest baseline \textbf{B}.}
\Description{Six scatter panels, one per dataset, plotting test PR-AUC against
end-to-end cost on a log scale. A grey line marks the Pareto front of each
panel and the letter B marks the strongest baseline, against which the
front's speed-up factors are given. The TCEE markers sit above and to the left
of both the structural baselines and the node-level fusion baselines in five of
the six panels.}
\label{fig:cost_quality}
\end{figure}

%% file: sections/6_discussion.tex
\section{Discussion}
\label{sec:discussion}

\looseness=-1
TCEE models both structure and time, and without a
backbone it beats every evaluated baseline on five of six AMLWorld
datasets in PR-AUC at lower end-to-end cost. We discuss the
implications and the next steps.

\paragraph{Deployment.} \looseness=-1 Transaction monitoring raises alerts for
investigators, and $90$ to $95\%$ of those from rule-based systems are
false positives~\cite{Oztas2024}. Investigators work through the alerts in
order, so a better ranking saves review time. Even on the low-illicit
datasets TCEE reaches up to $43.7$ PR-AUC, against $0.059$ for a random
ranking on the same test split. It can support an analyst, not replace one. Cost matters
because laundering methods change, requiring frequent
retraining~\cite{Oztas2024}. On the large graphs the full
TCEE-MLP finishes in about a day on one A40 GPU, where the structural
baselines need two to six days (Table~\ref{tab:cost}). How often banks
retrain is not public, so whether this suffices remains open.

\paragraph{Generalisation.} \looseness=-1 AMLWorld is generated to mirror real
transaction behaviour~\cite{Altman2023}. Whether these results hold on
real bank data, where labels are incomplete, should be explored.
AMLWorld also covers fixed typologies, most within two hops. TCEE-MLP
sees one hop, while the PNA backbone sees two and raises PR-AUC on the
four datasets trained to convergence.

\paragraph{Interpretability.} \looseness=-1 Each prediction exposes its
attention over both accounts' $F$ feature series and the $2w$ surrounding
transactions. An
analyst can point to this trace, whereas message-passing
detectors need separate explainers~\cite{li2024sefraud}. However, 
attention marks what the model consulted, not
why it decided~\cite{jain2019attention, wiegreffe2019attention}. We test whether
the neighbourhood attention weights are informative on HI-Small by deleting 
window transactions in order of their attention weight and measuring the drop in PR-AUC
(Figure~\ref{fig:window_deletion}). Removing the top $8$ transactions per account
causes over half of the drop from removing the whole window, against about a fifth
for $8$ random transactions.

\input{figures/window_deletion}

\paragraph{Limitations and Future Work.} \looseness=-1 These findings are limited
by one GPU, one frozen encoder, and one domain. Larger GPUs can hold the medium and
large embeddings resident, making TCEE faster. Future work can try other
TSFM encoders, or fine-tune one. TCEE
is also not tied to financial crime. It fits any multigraph whose nodes
carry a history of timed events, such as cybersecurity data.

%% file: figures/window_deletion.tex
\begin{figure}[tb]
\centering
\includegraphics{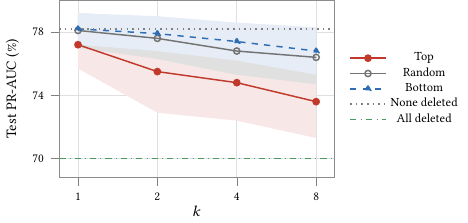}
\caption{Test PR-AUC on HI-Small after deleting the $k$ top, bottom or random
attended window transactions per endpoint.}
\Description{Line plot of test PR-AUC against the number of deleted window
transactions per endpoint, 1 to 8 on a log scale. Deleting the most attended
transactions lowers PR-AUC from 78.2 to 73.6 at eight, while deleting random
or least attended ones lowers it only to 76.4 and 76.8. Deleting the whole
window gives 70.0.}
\label{fig:window_deletion}
\end{figure}

%% file: sections/7_conclusion.tex
\section{Conclusion}
\label{sec:conclusion}

\looseness=-1
Financial crime detection requires understanding not only the
structure of financial networks, but also how the behaviour of
accounts evolves over time. Our work shows that combining
structural and temporal information can improve detection
considerably, and that the gains depend on how this information
is used in the model. When an account is represented the same way in all
of its transactions, those transactions can become hard to tell apart. TCEE
instead builds a separate representation for every transaction from its
own features, its accounts' histories and its local context. This design surpasses the strongest evaluated baseline on five of six anti-money laundering datasets while training $6.0\times$ faster.
Adding a GNN that captures broader structure yields further
gains on the smaller datasets, and the same combination also outperforms all
evaluated baselines on Ethereum phishing detection. Together, these results
suggest a simple lesson for detecting illicit transactions: bring the context directly 
to the point where the decision is made, the individual transaction.

%% file: sections/8_acknowledgements.tex
\begin{acks}
The support of the Swiss National Science Foundation (project number: 212158)
for this work is gratefully acknowledged. Research reported in this work was
partially facilitated by computational resources and support of the Delft AI
Cluster (DAIC) at TU Delft (RRID: SCR\_025091).
\end{acks}